# Developing an AI framework to automatically detect shared decision-making in patient-doctor conversations

**Authors:**

Oscar J. Ponce-Ponte[1,2,3]

David Toro-Tobon[2,3]

Luis F. Figueroa[3]

Michael Gionfriddo[2]

Megan Branda[2]

Victor M. Montori[2,3]

Saturnino Luz[4]

Juan P. Brito[2,3]

**Affiliations**:

[1]    Derriford Hospital, University Hospitals Plymouth NHS Trust, Plymouth, UK.

[2]    Knowledge and Evaluation Research Unit, Mayo Clinic, Rochester, MN, USA.

[3]    Care and AI lab, Knowledge and Evaluation Research (KER) Unit, Mayo Clinic, Rochester, MN, USA.

[4]    Centre for Medical Informatics, Usher Institute, University of Edinburgh, Edinburgh, UK.

## Abstract

**Background**: Shared decision-making (SDM) is necessary to achieve patient-centred care. Currently there is no methodology to automatically measure SDM at a large scale. This study aimed to develop an automated approach to measure SDM by using language modelling and the conversational alignment (CA) score.

**Methods**: A total 157 video-recorded patient-doctor conversations from a randomized multi-centre trial evaluating SDM decision aids for anticoagulation in atrial fibrillations were transcribed and segmented into 42559 sentences. Context-response pairs and negative sampling were employed to train deep learning (DL) models from scratch and fine-tuned BERT models of different sizes via the next sentence prediction (NSP) task. Each top-performing DL and BERT models were used to calculate four types of CA scores. Finally, we performed a random-effects analysis by clinician, adjusting for age, sex, race, and trial arm (decision aid vs. usual care) to assess the association between CA scores and SDM outcomes: the Decisional Conflict Scale (DCS) and the Observing Patient Involvement in Decision-Making 12 (OPTION12) scores. The *p* values were corrected for multiple comparisons with the Benjamini-Hochberg method.

**Results**: Among 157 patients (34% female, mean age 70 ±10.8), clinicians on average spoke more words than patients (1911 vs 773). The DL model without the stylebook strategy achieved a recall@1 of 0.227, while the fine-tuned $BERT_{BASE}$ (110M) achieved the highest recall@1 with 0.640. The AbsMax (18.36 SE ±7.74 *p*=0.025) and Max CA (21.02 SE ±7.63 *p*=0.012) scores generated with the DL without stylebook are associated with OPTION12. The Max CA score generated with the fine-tuned $BERT_{BASE}$ (110M) model is associated with the DCS score (-27.61 SE ±12.63 *p*=0.037). BERT model sizes did not have an impact on the association between CA scores and SDM measurements.

**Conclusion:** This study introduces an automated, scalable methodology to measure SDM in patient-doctor conversations through explainable CA scores. This approach has the potential for evaluating the effectiveness of implementing SDM strategies at scale.

1. **Introduction**

Shared decision-making (SDM) is a process in which doctors and patients take iterative and interactive steps to arrive at the best approach to address the patient's situation.[1] Despite the crucial role of SDM in achieving patient-centred care – a goal widely endorsed by regulators, payers, and healthcare guidelines[2,3] – SDM is not regularly assessed in practice. This is in part because many evaluation methods rely on self-reporting or require significant investments in time and expertise. Consequently, there is an urgent need to develop automatic, real-time, scalable measures to ensure that SDM is implemented effectively.

Modelling conversations by using natural language processing could form the basis for automation of SDM assessment. The development of such a model would first require the ability to understand the flow of discourse within patient-clinician interactions, accounting for the unstructured transitions between topics. Despite the recent advances in large language models (LLMs) for dialogue understanding,[4] current LLMs perform poorly in dialogue analysis and evaluation in terms of coherence, empathy, contradiction detection, and engagement.[5,6] Hence, to automatically measure SDM, a language model should be capable of analysing and evaluating conversations between patients and healthcare professionals.

Conversational alignment (CA), a psycholinguistic theory, has the potential to assist current language models in automatically measuring SDM between patients and clinicians. CA theory posits that a conversation is successful when the dialogue participants are in alignment with respect to their lexical choices, syntax, and semantics of the discourse as well as the alignment in the topic under discussion.[7] Similarly, SDM occurs when patients and clinicians jointly make healthcare decisions which requires mutual comprehension of the issues discussed throughout the consultation. Thus, we hypothesize that a dialogue model that incorporates the CA theory can be used to identify instances of SDM and the degree at which SDM has occurred in a conversation.

CA implies that participants have a tendency to unconsciously and automatically produce and interpret expressions in the same ways, meaning that CA is largely driven by the alignment at other levels of representation.[7] For instance, participants tend to produce similar words when the meaning of these words is aligned.[8] This can also be found in in their convergence in the use of expressions[9] or prosody.[10] The automatic measurement of CA has been performed by analysing the alignment at different levels of language,[11,12] and it has been found to be a practical method to measure social goals such as socialization skills in children with autism,[13] dialogue success,[5,11] decision-making tasks,[14] and empathy.[15] Therefore, a method that incorporates CA could automate the detection of instances of SDM, providing a more efficient alternative to current techniques that rely on human involvement.

There have been deep learning (DL)-based models that employed the transformers architecture to measure CA.[13,16] However, to our knowledge the methods to measure CA have not yet leveraged the capabilities of pre-trained LLMs and there are no currently validated CA-based dialogue models that can identify instances of SDM. Therefore, our objective is to develop an automated approach to measuring SDM from transcripts of patient-doctor conversations using language modelling and adding an intermediate, explainable step; the conversational alignment (CA).

## 2. Methods

This study follows a three-step approach **(Figure 1)**. First, we developed DL models able to distinguish different conversational patterns or styles between patients and clinicians. Second, we employed these trained models to measure CA. Finally, we validated our results against two of the most common SDM measurement tools: the OPTION12[31] and the Decisional Conflict Scale (DCS).[32]

**Figure 1**: Overall development and validation approach

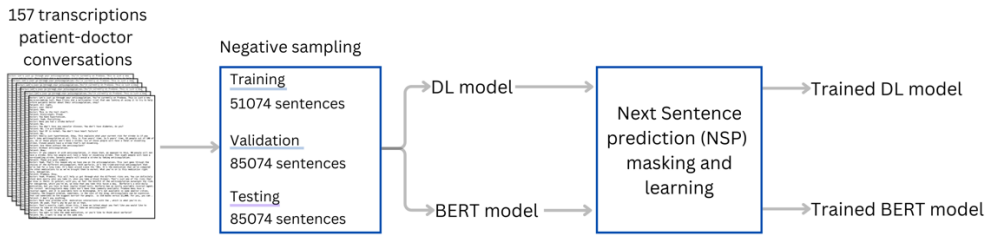
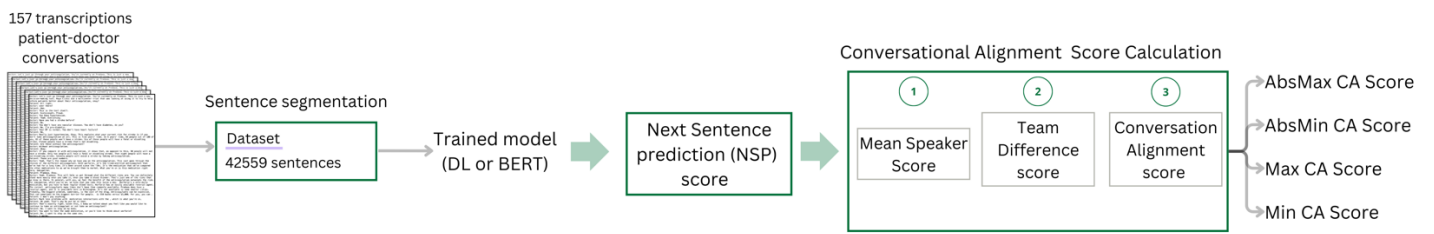
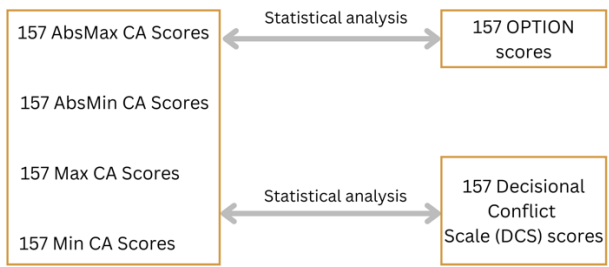

We used video-recorded conversations from a randomized multi-centre trial that evaluated the impact of using a decision aid to facilitate SDM around anticoagulation for patients with atrial fibrillation (AFib), compared against usual care.[15] We included a total of 157 manually transcribed conversations that had the DCS and the OPTION12 scores.[31,32] Conversations where a third person (e.g., another healthcare professional, relative, or carer) participated were excluded. All transcripts were reviewed by a trained annotator who removed transcribed paralinguistic features (e.g., laugh, pause). Contractions such as ['cause] or [AFib] were preserved because they reflect spoken styles and could help capture alignment in conversational style.[33,34] However, contractions that depend on the listeners' perception (i.e., the transcriber) were normalised (e.g., [goin'] was changed to [going]).[35] Abbreviations with pronunciations consistent with other/elongated forms were normalised (e.g., p.m. to PM or St. to Saint). All punctuation marks were preserved. This study was approved by the institutional review board (IRB) at Mayo Clinic, USA (IRB#23-002499).

*2.1.   Language modelling*

We trained DL models from scratch, and fine-tuned a pretrained LLM, the BERT (Bidirectional Encoder Representations from Transformer) model.[27] These models are trained with sentences rather than full turns. While a turn may contain multiple sentences, focusing on individual sentences helps the model identify distinct intents and functions within and across turns.[25] This approach increases the sample size for dialogue modelling while reducing input length. For training, we used the next sentence prediction (NSP) task, which enables the model to learn conversational style, discourse structure, and semantic variations. [36,37] In NSP, sentences are masked, and the model predicts such sentences. This approach, known as self-supervised learning (SSL), is claimed to be similar to human learning, as it requires the model to infer information from context. [22,38,39]

We defined our NSP task as follows; let $D$ be a dialogue or conversation within our dataset and $T$ the number of sentences such that $D = (s_1, s_2, s_3, ..., s_T)$, where context is $C = (s_1, s_2, s_3, ..., s_{T-1})$ and response is $R = (s_T)$. Context ($C$) was generated by concatenating five continuous sentences, and the ground truth response sentence ($R$) was the following sentence (i.e., sixth sentence). These ($C,R$) pairs are called positive samples and all 157 transcriptions were divided into 42559 positive samples. These were divided into 60%, 20%, and 20% for the training, validation and testing, resulting in 25537, 8511, and 8511 positive pairs for each dataset respectively. Because the language models follow a binary classification task, for each positive pair, we randomly sampled false responses from within the same dataset where positive responses were previously excluded, resulting in negative sample pairs. We created one negative sample per positive sample in the training dataset, and nine negative samples per positive sample for both the validation and test dataset.[13] This approach resulted in a total of 51074, 85110, and 85110 samples for the training, validation, and test datasets, respectively. The model's objective for each ($C,R$) was to determine the probability $p(R|C)$ of the response being ground truth given the context, approximated by the model as $y \in [0,1]$. The loss function was cross-entropy.

*2.1.1. Deep Learning and BERT models*

The DL and BERT models were tested with the recall@K performance metric,[40] and only the model with the best performance within each group were selected for the generation of CA scores and validation against SDM outcomes. Recall@k is one of the main performance metrics for language understanding models.[40,41] It measures the proportion of true positive responses among the top-*k* selected responses from the list of *n* available candidates for one context. The available candidates for our training dataset are 2 (*n*=2) and for the evaluation and testing datasets are 10 (*n*=10). We employed a recall@1 for the training and validation stages, and recall@1, recall@2, and recall@5 for testing.

The DL models were developed inspired by Yu *et al.'s* work.[13] We trialled several modifications to their architecture that were tested on the validation dataset. Our architectural modifications to the model are represented with dashed boxes on **Figure S1**. A detailed description of the deep learning models is available in the supplementary material. Briefly, the two DL models follow an Encoder-Decoder framework and were built based on the self-attention mechanism[42] as well as the long short-term memory (LSTM) architecture.[43] The only difference between these two models was the presence or absence of the stylebook method **(Figure S2)** which is implemented with the cross-attention mechanism by initialising the keys (*K*) and values (*V*) of the model with random weights. The stylebook method is inspired by the global style tokens strategy commonly used in Text-to-Speech systems which capture speaker characteristics and are capable of transferring these styles to synthesised speech.[44] This stylebook methodology is meant to capture and differentiate each participant's style from both patients and doctors.

The BERT model is a pre-trained LLM that has good performance in dialogue understanding tasks.[19,27] We added an NSP classification head on top of the BERT model. **(Figure S3)** The $BERT_{BASE}$ model without fine-tuning and the fine-tuned $BERT_{BASE}$ model were selected for our main analysis. The $BERT_{BASE}$ has a total of 110 million (110M) parameters. We further explored the effect of models with different number of parameters on the performance in dialogue understanding. These include $BERT_{LARGE}$ (335M) the $BERT_{MEDIUM}$ (41.4M), $BERT_{SMALL}$ (28.8M), and $BERT_{TINY}$ (4.4M). Additionally, we explored the impact of inserting the [SEP] token in between sentences within the context for each model size which may help the model understand the separation between sentences within the context.[45] A detailed description of this methodology is available in the supplementary material.

*2.1.2. Implementation and reproducibility*

All models were developed with Pytorch[46] and Lightning[46]. We employed the AdamW optimizer[47] and 16-mixed precision[48] for training in 1 to 16 Tesla V100-SXM2-16GB GPUs. Each model's training characteristics and hyperparameter tuning results are available on **Tables S5 to S9**. The coding for our models as well as training, validation, and testing features are available upon request.

*2.2. Conversational alignment*

To generate the CA scores, we first processed the transcripts by segmenting each conversation into pairs of context and response (*C,R*) but in this case no negative sampling was performed. We then used our best DL and BERT pre-trained dialogue models at inference time to calculate the probability of (*R*) following the context (*C*), or *P(R|C)*.

To calculate the CA score, we divided each conversation as equally as possible into 10 intervals based on the number of tokens without breaking up individual sentences. Only intervals that had sentences from both the doctor and the patient were included. In each of the 10 intervals, all sentences predicted probabilities (i.e., *P(R|C)*) of a given speaker were added up and divided by the total number of sentences by the same speaker within that interval. This resulted in two mean scores per interval: one for the doctor and one for a patient. Within each interval, we used each speaker's mean score and calculated the team difference score. This validated approach on textual[23] and acoustic data[49] measures how different the styles of the participants are within each interval, resulting in a team difference score per interval and 10 team difference scores per conversation. We then calculated the temporal convergence of the 10 similarity scores per conversation or the CA score.[13] Because convergence can result in four different types of CA scores, we tested four different CA scores per conversation. The supplementary material has a detailed description of this method.

### 2.3. Shared decision-making validation

#### 2.3.1. Scores description

We assessed the association between the four generated CA scores per model with the DCS[32] and the OPTION12 scale.[31] The DCS is a self-administered questionnaire that reflects the patient's uncertainty in the decision being made. This score can range from 0 to 100, with higher scores indicating greater decisional conflict.[32] The OPTION12 scale is an observed encounter outcome where conversations were scored by trained assessors who judged whether the clinician actively involved patients in the decision-making of their healthcare plans.[31] This score also ranges from 0 to 100, where higher scores indicate higher effort to involve patients. The interrater reliability for assessors ranged from 0.84% to 0.96%.[50]

#### 2.3.2. Statistical analysis

After normalising alignment scores that did not have a normal distribution, we implemented generalised linear models (GLM) to perform an unadjusted analysis between the CA scores and the DCS as well as OPTION12 scale. We then performed a GLM multivariable analysis adjusted for age, sex (male and female), race (Asian, Black, White, Native-American, and multiple), and the group they were assigned to during the clinical trial (decision aid vs. no decision aid). Finally, because some doctors participated in more than one patient-doctor conversation, we performed a random-effects analysis by clinician.[51] All of the results are presented with their estimates, corresponding standard errors, and their *p*-values.

We had four different CA scores per model (one DL model and one BERT model), and each of them was analysed with both of our outcomes (DCS and OPTION12), resulting in a total of 16 comparisons. Considering the number of comparisons, we performed a Benjamini-Hochberg *p*-value correction to the the random-effects analysis. We assumed a false discovery rate (FDR) of 20%.[52] As part of our exploratory analysis we tested whether CA scores generated by different sizes of the BERT model would have a stronger association with SDM outcomes. All statistical analyses were performed with the SAS software.

3. **Results**

Out of 922 encounters in the trial, we identified 157 eligible patient-doctor conversations comprising a set of 42559 sentences. Patients, on average, spoke around 773 words over a mean of 98 sentences, and doctors spoke an average of 1911 words over a mean of 161 sentences. There were only 10 conversations in which patients spoke more words than their doctors. The mean duration of the conversations was 32 minutes (SD ± 16). Patients' and clinicians' demographics are shown in Tables 1 and 2.

**Table 1**: Patients baseline characteristics

|  | Overall (n= 157) |
|---|---|
| **Age, Mean (SD)** | 70.15 (10.8) |
| **Sex, n (%)** |  |
| Female | 53 (33.8) |
| Male | 104 (66.2) |
| **Patient Race, n (%)** |  |
| White | 125 (79.6) |
| Black | 22 (14.0) |
| AI/AN | 1 (0.6) |
| Asian | 2 (1.2) |
| Multiple | 4 (2.5) |
| Missing | 3 (1.9) |
| **Education Level, n (%)** |  |
| Highschool or less | 35 (22.3) |
| Some College or Technical degree | 48 (30.6) |
| 4 year College Degree | 31 (19.7) |
| Graduate degree | 27 (17.2) |
| Missing | 6 (3.8) |
| **Inadequate health literacy, n (%)** |  |
| Inadequate health literacy | 15 (9.6) |
| Adequate health literacy | 139 (88.5) |
| Missing | 3 (1.9) |
| **Duration of Encounter (minutes), Mean (SD)** | 31.9 (16.1) |
| **CHADS-VASC Score, Mean (SD)** | 3.3 (1.5) |

**Table 2**: Clinicians baseline characteristics

|  | Total (N=59) |
|---|---|
| **Age (years), Mean (SD)** | 46.5 (13.4) |
| **Sex, n (%)** | |
|   Female | 25 (43.9%) |
|   Male | 32 (56.1%) |
|   Missing | 2 |
| **Clinician type, n (%)** | |
|   Nurse practitioner | 7 (12.3%) |
|   PA | 1 (1.8%) |
|   Physician | 42 (73.7%) |
|   Pharmacist | 3 (5.3%) |
|   RN | 4 (7.0%) |
|   Missing | 2 |
| **Practice type, n (%)** | |
|   Family Medicine | 7 (12.3%) |
|   Internal Medicine | 13 (22.8%) |
|   Cardiology | 11 (19.3%) |
|   Cardiac Electrophysiology | 17 (29.8%) |
|   Other | 9 (15.8%) |
|   Missing | 2 |

*3.1. Language modelling*

Overall, the DL model without stylebook had better performance than the DL model with stylebook by around 0.03 in Recall@1, Recall@2 and Recall@5. The $BERT_{BASE}$ fine-tuned with patient-doctor conversations outperformed the $BERT_{BASE}$ without fine-tuning by 0.2, and the fine-tuned model had the highest performance compared to the other DL models. The most expensive model to fine-tune is the $BERT_{BASE}$ in terms of model size and GPU usage **(Table 3)**.

Our exploratory analysis shows that fine-tuning BERT models with a higher number of parameters results in better performance **(Table 3)**. However, using the [SEP] token to separate the sentences to model contextual information results in negligible differences in performance **(Table S10)**.

**Table 3:** Recall@k metrics of language models in language understanding of patient-doctor conversations

| Model name | Size | GPU | Recall@1 | Recall@2 | Recall@5 |
|---|---|---|---|---|---|
| **Primary Analysis** | | | | | |
| DL without stylebook | 11.1M | 1 | 0.227 | 0.390 | 0.715 |
| DL with stylebook | 20.8M | 1 | 0.187 | 0.331 | 0.668 |
| $BERT_{BASE}$ without fine-tuning | 110M | * | 0.439 | 0.584 | 0.838 |
| $BERT_{BASE}$ with fine-tuning | 110M | 4 | 0.640 | 0.792 | 0.942 |
| **Exploratory Analysis** | | | | | |
| $BERT_{LARGE}$ | 335M | 8 | 0.635 | 0.789 | 0.945 |
| $BERT_{BASE}$ | 110M | 4 | 0.640 | 0.792 | 0.942 |
| $BERT_{MEDIUM}$ | 41.4M | 1 | 0.568 | 0.731 | 0.915 |
| $BERT_{SMALL}$ | 28.8M | 1 | 0.552 | 0.711 | 0.908 |
| $BERT_{TINY}$ | 4.4M | 1 | 0.471 | 0.625 | 0.856 |

DL: Deep Learning
* No GPUs were used because $BERT_{BASE}$ did not require fine-tuning

## *3.2. Conversational alignment*

Unadjusted and adjusted analyses show that the AbsMax and Max CA scores calculated with the DL without stylebook language model is associated with the OPTION12 score. Similarly, the CA Max score calculated with the fine-tuned $BERT_{BASE}$ model is associated with the DCS measurement. Because the ratio of patients to doctors was 3:1, we performed a mixed-effects analysis by clinicians confirming the results found in the adjusted analysis. These results remained consistent even after adjusting for multiple comparisons **(Table 4)**.

**Table 4:** Random-effects multivariable analysis of Conversational Alignment Scores by Outcome and Model

| Outcome | Model | CA Method | Unadjusted analysis Estimate (±SE) | p value | Adjusted analysis* Estimate (±SE) | p value | Mixed-effects† Estimate (±SE) | p value |
|---|---|---|---|---|---|---|---|---|
| OPTION12 | BERT$_{BASE}$ 110M | AbsMax | 0.18 (±11.03) | 0.986 | -2.48 (±11.44) | 0.828 | -3.3 (±10.89) | 0.762 |
| | | AbsMin | -0.27 (±0.86) | 0.753 | 0.33 (±0.89) | 0.712 | 0.08 (±0.83) | 0.920 |
| | | Max | 2.24 (±11.39) | 0.844 | -0.83 (±11.95) | 0.944 | -1.30 (±11.20) | 0.908 |
| | | Min | -61.40 (±59.67) | 0.305 | -49.94 (±59.16) | 0.400 | -38.33 (±48.93) | 0.436 |
| | DL without Stylebook | **AbsMax** | **23.05 (±7.98)** | **0.004** | **21.40 (±8.65)** | **0.014** | **18.36 (±7.74)** | **0.020¶** |
| | | AbsMin | 1.00 (±0.76) | 0.190 | 1.02 (±0.76) | 0.184 | 0.90 (±0.68) | 0.186 |
| | | **Max** | **25.27 (±7.97)** | **0.001** | **22.90 (±8.48)** | **0.007** | **21.01 (±7.63)** | **0.007¶** |
| | | Min | 10.64 (±44.05) | 0.809 | 18.14 (±43.80) | 0.679 | 7.48 (±38.81) | 0.848 |
| DCS | BERT$_{BASE}$ 110M | AbsMax | -18.81 (±11.99) | 0.119 | -23.16 (±12.49) | 0.065 | -21.68 (±12.31) | 0.082 |
| | | AbsMin | -0.17 (±0.96) | 0.855 | 0.33 (±1.00) | 0.735 | 0.49 (±0.98) | 0.618 |
| | | **Max** | **-24.35 (±12.24)** | **0.048** | **-29.75 (±12.85)** | **0.022** | **-27.6 (±12.63)** | **0.032¶** |
| | | Min | 20.59 (±64.59) | 0.750 | 25.85 (±64.10) | 0.687 | 32.84 (±61.09) | 0.592 |
| | DL without Stylebook | AbsMax | 1.44 (±8.99) | 0.873 | 3.90 (±9.77) | 0.690 | 4.74 (±9.43) | 0.617 |
| | | AbsMin | 0.99 (±0.84) | 0.243 | 0.82 (±0.85) | 0.337 | 0.89 (±0.82) | 0.283 |
| | | Max | 2.41 (±9.00) | 0.789 | 4.94 (±9.57) | 0.606 | 5.39 (±9.28) | 0.563 |
| | | Min | 27.85 (±47.87) | 0.562 | 32.70 (±47.54) | 0.493 | 29.92 (±45.92) | 0.517 |

CA: Conversational alignment, SE: Standard Error, DL: Deep Learning

* Adjusted for age, sex, race, and clinical trial arm (intervention vs. control).

† Mixed-effects for clinician and adjusted for age, sex, race, and clinical trial arm (intervention vs. control).

¶ After adjusting for multiple comparisons with Benjamini-Hochber and false discovery (FDR) rate of 20%

Our exploratory analysis shows that when calculating CA scores with different sizes of the fine-tuned BERT model, the Max CA scores calculated with models that are above 55 million parameters (BERT$_{LARGE}$ - 330M, BERT$_{BASE}$ - 110M, and BERT$_{MEDIUM}$ 55M) are consistently associated with the DCS measurement. However, unlike the AbsMax CA score calculated with the BERT$_{BASE}$ - 110M model which was found to not be associated with the DCS, the AbsMax CA scores calculated with BERT$_{LARGE}$ 330M and BERT$_{MEDIUM}$ - 55M were associated with the DCS (Table S13).

## 4. Discussion

We developed and validated a new automated method to measure SDM by measuring CA in conversations between patients and doctors and by modelling these conversations with a pre-trained LLM and a transformers-based architecture. We found that, on average, for every unit increase in the AbsMax and Max CA scores, derived with the DL without stylebook model, is associated with an increase of 18.36 (±SE 7.74) and 21.01 (±SE 7.63) points in the OPTION12 score respectively, and a one-unit increase in the Max CA score, derived with the BERT model, is associated with a reduction of 27.61 points in the DCS score (-27.61 ±SE 12.63).

Our automatic SDM measurement methodology introduces an intermediate step - the CA score calculation from Figure 1. The benefit of this is that the conversation can be explored to understand which part was the most important for score calculation. For instance, supplementary figures S4 and S5 show parts of two conversations considered the most important by the CA calculation methodology. This granular analysis offers actionable feedback to healthcare professionals by highlighting parts of the conversation that enhance or hinder CA and, in consequence, shared decision-making.

The automatic measurement of SDM largely relies on language modelling. While most current language models use word-level prediction for training,[17] using the NSP enables an understanding of different styles and discourses of language which is critical to calculate the CA.[18–22] NSP seems to enhance the model's ability to emulate human understanding and process contextual details,[22] aligning with our methodology for CA score calculation,[13,23] which depends on context and changes of style that occur within a conversation.[24] Prioritizing context through the NSP task and CA is vital for patient-doctor interactions, where distinct language styles and individual agendas shape the dynamic of conversations.

One of the main differences between our work compared to other dialogue modelling methods[13] is that we used sentences as unit of analysis instead of dialogue turns. A conversation is a turn-taking event where each turn ends when a specific amount of pause time occurs. Unlike turns, using sentences as the unit of analysis captures different conversational functions within a turn.[18,25] This is especially important for patient-doctor conversations where the turns can contain multiple sentences because both actors tend to give more explanatory statements than short, direct questions.[26]

Additionally, our exploratory analysis showed that the size of pretrained LLMs does not influence CA score calculation or its association to SDM measurements. While the scaling law states that performance improves as models increase in size,[27–30] our findings deviate in the context of CA. For instance, CA scores generated with $BERT_{MEDIUM}$ 55M outperformed $BERT_{LARGE}$ - 335M and $BERT_{TINY}$ - 4.4M models.

One of the limitations of our study may be the limited sample size of 157 transcribed patient-doctor conversations. However, this study benefitted from choosing sentences as unit of analysis as this increased the sample size to 42559 sentences. Additionally, even though the association of CA scores with SDM measurements was validated on only 157 conversations, we found positive results even after mixed-effects analysis and correcting for multiple comparisons. Another limitation is that the CA methodology might underestimate alignment in long-term patient-doctor relationships, where pre-existing convergence in language styles could be misclassified as low alignment in this study. Future studies should account for duration and quality of patient-doctor pre-existing relationships. Finally, our method was evaluated only on dyadic interactions, in consequence its efficacy in conversations involving additional participants (e.g., caregivers, family member) remains unknown.

This is the first study that has developed a fully automated approach to measure SDM in patient-doctor conversations. It introduces an intermediate, explainable approach which allows healthcare actors as well as researchers identify where the calculation came from. Our methodology has the potential to aid in measuring the effectiveness of implementing SDM strategies at large scale. In future work, we will validate this method in multiparty healthcare conversations and in multilingual settings, as well as to improve the dialogue model by including non-linguistic features of language such as visual features (i.e., gestures), and acoustic-prosodic features.

## 5. Author contribution

O.J.P-P and S.L. designed the study. J.P.B and V.M acquired the data. O.J.P-P developed, trained, and tested the deep learning models. M.B. performed the statistical analysis. O.J.P-P drafted the article. J.P.B, S.L, D.T-T, M.G, and L.F critically revised the manuscript. All authors reviewed, edited, and approved the final version of the manuscript for submission.

## 6. Competing interests

The authors declare no competing interests.

# Supplementary material

**Table of Contents**

# 1. Methods

## 1.1. Deep learning models architecture

Both the DL model with stylebook and the DL model without stylebook follow an Encoder-Decoder (EC) framework. The only difference between them is how the Keys (*K*) and Values (*V*) are initialised - the DL model stylebook model initialise them with random weights. (**Figure S1)** The model shares the encoding architecture for both the context and the response. This architecture contains a transformer layer followed by an Long Short-Term Memory (LSTM) layer.[1] After this, both representations go through the decoder network which follows a matching-aggregation framework[2] that includes only the self-attention mechanism from a transformer block[3] followed by a LSTM layer.

**Figure S1**: Deep Learning with Stylebook model architecture

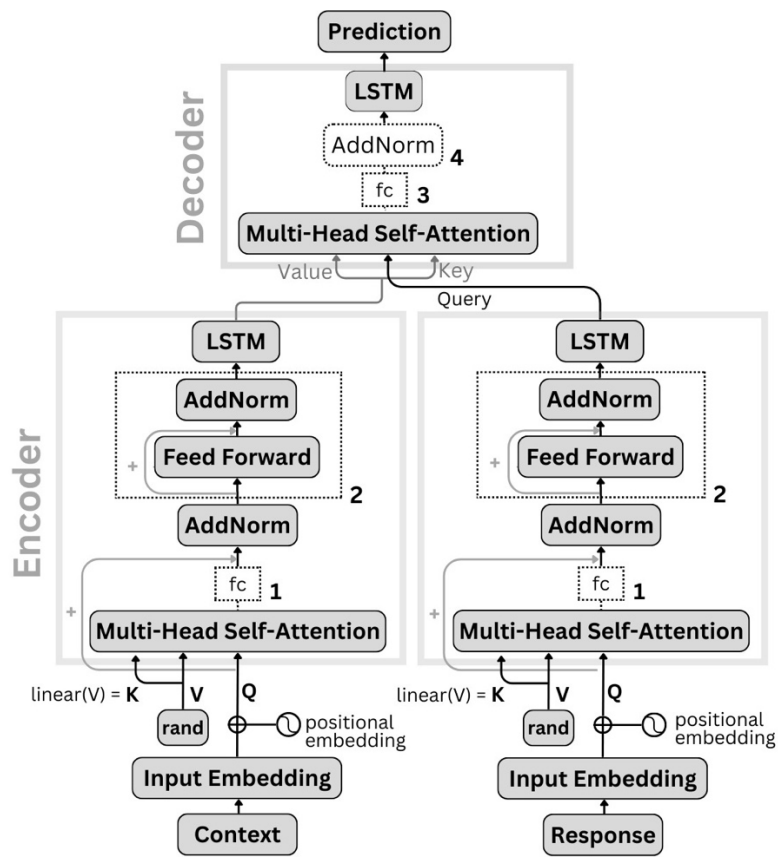

*Four dashed boxes = New architectures that were tested on the validation set.*
*rand = random numbers with normal distribution, linear = linear transformation, fc = fully-connected layer, AddNorm = residual connection and a normalization layer, Feed Forward = Feed Forward layer, LSTM = Long Short-Term Memory*

*i     Text encoding*

**Subword and Positional embedding**: The first step of the architecture is the embedding of words and positions. The text input is converted into subword embeddings by using BPEmb, a model pre-trained on Wikipedia by using Byte-Pair Encoding (BPE).[4] We implemented the BPEmb subword embedding with 300 dimensions and a vocabulary of 100,000. This process transforms both context and response into *n* subword embeddings $s = (s_{w1}, s_{w2}, ..., s_{wn})$. Now *s* is in $R^{n \times d}$, where *n* is the token length of the context $s_c$ or response $s_r$, and *d* is the embedding size (300). After this, positional embeddings are added to the subword embeddings.

*ii    Encoder*

**Transformer Encoder**: The transformer encoder layer is shared for both the context and response. Wee tested two different modifications in this section: 1) the addition of a fully-connected (fc) layer, and 2) the addition of a Feed Forward and a normalisation layer with a residual connection (AddNorm). These modifications are inspired by a study that found that increasing the number of normalisation layers (AddNorm) can have a positive impact on the performance of transformers. Another study showed that adding and increasing the dropout rate in the fc layer can be beneficial,[6] which informed the option of not using the fc layer. These assumptions were tested in the validation set.

In both models (with or without the stylebook), the input embedded representation or query ($Q$) goes through a multi-head self-attention layer. In the stylebook model, the key($K$) and value($V$) are global for all $Q$. The $V$ is initialized with random weights and the $K$ is a linear transformation of $V$. Key-value pairs are global and shared throughout training. In the model without stylebook, the key($K$) and value($V$) are the same as the query($Q$). After this, the output of this multi-head self-attention goes through an fc layer, a normalisation layer with a residual connection (AddNorm), a feed forward layer, and a second AddNorm. The output of this transformer block is a matrix $\in R^{n \times d}$.

**LSTM Encoder**: The LSTM layer is also shared for both context and response. LSTM has the ability to learn short and long-term dependencies.[1] The input for this layer is the output of the transformer layer and the output $\in R^{n \times h}$ where $n$ is the context and response token length and $h$ is the number of hidden units (set to a specific number).

*iii    Decoder*

The decoder performs a discriminative task where the outputs are predicted probabilities that represent how likely is that the response sentence is a false response or ground truth response.

**Self-attention**: This section was set to one head. The query $Q$ is the response encoding output $\in R^{n_r \times h}$, and key-value pairs are the context encoding output $\in R^{n_c \times h}$ where $h$ is the number of hidden units. The self-attention mechanism is also known as cross-attention and is widely used in machine translation tasks. At the beginning of this layer, we applied a linear transformation of the three inputs (*Q*, *K*, and *V*) converting the last dimension (*h*) into the embedding size dimension (*d*). These new three inputs were put into the model and the output was a matrix of $R^{n_r \times d}$ where $n_r$ is the token length of the response input. There are two additional architectures that we tested on the validation set: 1) the addition of the fc layer, and 2) the addition of AddNorm.

**Aggregation decoder and prediction**: This aggregates the matching, or output, of the previous layer by using LSTM with a single layer. The output of the LSTM has the dimensions $R^{1 \times h}$, where *h* is the dimension of the hidden size of the aggregation decoder which is put into a dense layer (output $\in R^{1 \times 2}$). The output then goes into a softmax function that generates the output probability of being a false or ground truth response, namely *y* ∈ [0,1].

## 1.2. The BERT models architecture

The BERT (Bidirectional Encoder Representations from Transformers) model is a pretrained large language model (LLM) used in dialogue models that achieved SOTA results in language understanding tasks.[7,8] For this study, we fine-tuned the BERT model by adding an NSP classification head on top. **(Figure S2)** Similarly, to the previous section (Models 1 and 2), the context sentences are concatenated into a long sentence.

**Figure S2**: Model 3 (BERT model) architecture

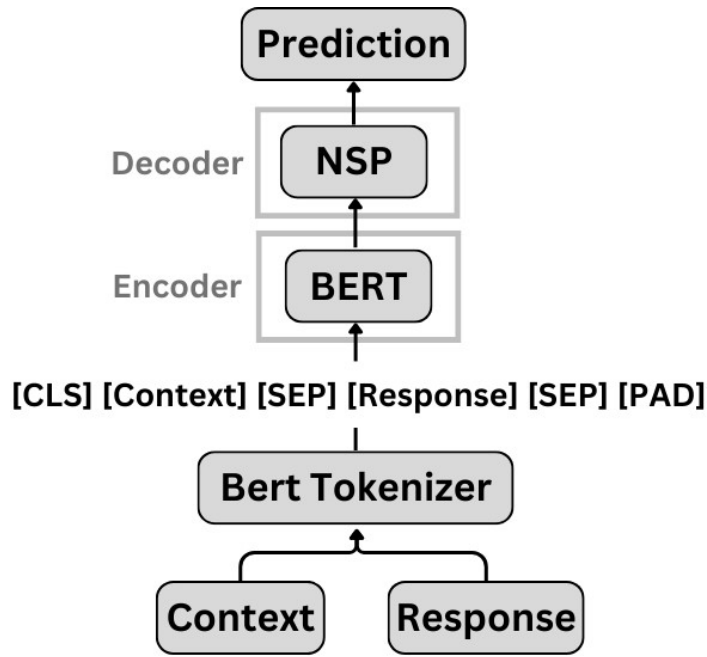

*i     Encoder*

First the text from both the context and the response is fed into the BERT tokenizer where the input is WordPiece-tokenized.[8] This introduces two additional tokens: 1) [CLS]: this is inserted as the first token of the whole sequence and captures the information of the whole input (context and response), which is used for classification tasks (NSP), and 2) [SEP]: this token is added in between the context and response, and at the end of
response.

$$EC = Encoder([CLS]\ Context\ [SEP]\ Response\ [SEP])$$

*ii     Decoder*

A classification head is assembled on top of the encoder layer. This is the "next sentence prediction" (NSP) head that will generate two logit values that correspond to the ground truth response and false response. We transformed these logits into probabilities by using a softmax function.

$$Y = Decoder(EC)$$

*iii    Comparing different BERT model sizes*

Both the BERT$_{BASE}$ and the BERT$_{LARGE}$ were described in the original BERT article[7]. Smaller sizes for the BERT model including BERT$_{TINY}$, BERT$_{SMALL}$, and BERT$_{MEDIUM}$, were later released by Turc *et al.*[9] **(Table S1)**.

**Table S1**: Characteristics of BERT models with different number of parameters

| Model | Total parameters | Number of layers* | Hidden size | Self-attention heads | Feed-forward/filter size |
|---|---|---|---|---|---|
| BERT$_{TINY}$ | 4.4M | 2 | 128 | 2 | 512 |
| BERT$_{SMALL}$ | 28.8M | 4 | 512 | 8 | 2048 |
| BERT$_{MEDIUM}$ | 41.4M | 8 | 512 | 8 | 2048 |
| BERT$_{BASE}$ | 110 million | 12 | 768 | 12 | 3072 |
| BERT$_{LARGE}$ | 340 million | 24 | 1024 | 16 | 4096 |

* Transformer blocks

*1.3.    Hyperparameter tuning*

All models were developed with Pytorch[10] and Lightning.[11] For all of our models, we used the AdamW optimizer[12] and 16-mixed precision[13] for training. All models were trained by using from 1 to 16 Tesla V100-SXM2-16GB GPUs. To be comparable to the BERT's model, the maximum token length for all models was set to 512.

*i    Deep learning models*

We tuned all the hyperparameters of the model for the DL model with stylebook and used these parameters to train the DL model without stylebook. Hyperparameter tuning was performed on the validation dataset. Our strategy was to first use early stopping with a patience of 10 and find the minimum recall value.[14] In the DL model with stylebook, the *V* is initialised with random numbers from a normal distribution with mean 0 and variance 1, and the key(*K*) was a linear transformation of *V*. In the model without stylebook, value (*V*) and key (*K*) are the same as the query (*Q*).

The initial hyperparameters were the ones found to be best by Yu et al.'s study.[15] The embedding dimension was 300, the stylebook size 500, the number of heads of the encoder multi-head self-attention was four, the number of forward expansions was four, the LSTM encoder had one layer and 512 hidden units, the aggregation LSTM had one layer and 128 hidden units, and dropout was set to zero. In terms of the architecture, there are four different new modifications to the architecture: 1) the addition of a fully-connected layer (FC1), 2) addition of a Feed Forward layer (two linear transformations layers and the activation function called Gaussian Error Linear Unit or GELU[16]) and an AddNorm (FF & AddNorm1), 3) the addition of a second fully-connected layer (FC2), and 4) the addition of AddNorm (AddNorm2). The first two modifications belonged to the encoder part of the model and the last two to the decoder part. We initialised the hyperparameter tuning with the FC1 but without the FF&AddNorm1, FC2, and AddNorm2. We first tuned batch sizes from 16 to 128 and a learning rate from $1\times10^{-2}$ to $1\times10^{-5}$. **(Table S2)**

**Table S2:** Recall@1 results on the validation dataset for batch size and learning rate.

| Batch Size | Learning rate | Last epoch | Recall@1 |
|---|---|---|---|
| 8 | $1\times10^{-2}$ | 11 | 0 |
| 8 | $1\times10^{-3}$ | 11 | 0 |
| 8 | $1\times10^{-4}$ | 12 | 0 |
| 8 | $1\times10^{-5}$ | 13 | 0.073 |
| 16 | $1\times10^{-2}$ | 11 | 0 |
| 16 | $1\times10^{-3}$ | 11 | 0 |
| 16 | $1\times10^{-4}$ | 18 | 0.028 |
| 16 | $1\times10^{-5}$ | 12 | 0.095 |
| 32 | $1\times10^{-2}$ | 11 | 0 |
| 32 | $1\times10^{-3}$ | 11 | 0 |
| 32 | $1\times10^{-4}$ | 12 | 0.055 |
| **32** | **$1\times10^{-5}$** | **21** | **0.150** |
| 64 | $1\times10^{-2}$ | 11 | 0 |
| 64 | $1\times10^{-3}$ | 14 | 0 |
| 64 | $1\times10^{-4}$ | 12 | 0.100 |
| 64 | $1\times10^{-5}$ | 17 | 0.103 |
| 128 | $1\times10^{-2}$ | 11 | 0 |
| 128 | $1\times10^{-3}$ | 17 | 0 |
| 128 | $1\times10^{-4}$ | 24 | 0.112 |
| 128 | $1\times10^{-5}$ | 28 | 0.088 |

After selecting the batch size of 32 and a learning rate of $1 \times 10^{-5}$, we tune the four modifications of the model: FC1, FF & AddNorm1, FC2, and AddNorm2. **(Table S3)**

**Table S3:** Recall@1 results on the validation dataset for the four modifications of the model

| FC1 | FF & AddNorm1 | FC2 | AddNorm2 | Last epoch | Recall@1 |
|---|---|---|---|---|---|
| Yes | No | No | No | 21 | 0.150 |
| Yes | No | Yes | Yes | 11 | 0.120 |
| Yes | No | No | Yes | 16 | 0.116 |
| Yes | No | Yes | No | 19 | 0.134 |
| Yes | Yes | No | No | 16 | 0.049 |
| Yes | Yes | Yes | Yes | 11 | 0.122 |
| Yes | Yes | No | Yes | 18 | 0.112 |
| Yes | Yes | Yes | No | 28 | 0.129 |
| No | No | No | No | 17 | 0.072 |
| No | No | Yes | Yes | 50 | 0.175 |
| No | No | No | Yes | 14 | 0.125 |
| No | No | Yes | No | 18 | 0.039 |
| No | Yes | No | No | 31 | 0.080 |
| **No** | **Yes** | **Yes** | **Yes** | **50** | **0.176** |
| No | Yes | No | Yes | 17 | 0.120 |
| No | Yes | Yes | No | 26 | 0.001 |

The best architecture for our model was the one without the FC1 but with FF & AddNorm1, FC2, and AddNorm2 achieving a Recall@1 of 0.176. We then continued tuning the remaining hyperparameters as shown in Table S4 and selected the best model to be tested on the testing dataset.

**Table S4**: Recall@1 results on the validation dataset for the remaining features of the architecture

| Number of heads | Number of forward expansions | Number of encoding | LSTM encoder | LSTM aggregation hidden size | Dropout | Last epoch | Recall@1 |
|---|---|---|---|---|---|---|---|
| 4 | 1 | 1 | 512 | 128 | 0 | 50 | 0.176 |
| 6 | 1 | 1 | 512 | 128 | 0 | 50 | 0.173 |
| 5 | 1 | 1 | 512 | 128 | 0 | 50 | 0.173 |
| 3 | 1 | 1 | 512 | 128 | 0 | 50 | 0.177 |
| 2 | 1 | 1 | 512 | 128 | 0 | 12 | 0.122 |
| 1 | 1 | 1 | 512 | 128 | 0 | 50 | 0.174 |
| 3 | 2 | 1 | 512 | 128 | 0 | 49 | 0.174 |
| 3 | 3 | 1 | 512 | 128 | 0 | 44 | 0.171 |
| 3 | 4 | 1 | 512 | 128 | 0 | 48 | 0.174 |
| 3 | 1 | 2 | 512 | 128 | 0 | 50 | 0.180 |
| 3 | 1 | 3 | 512 | 128 | 0 | 50 | 0.172 |
| 3 | 1 | 4 | 512 | 128 | 0 | 50 | 0.178 |
| 3 | 1 | 5 | 512 | 128 | 0 | 50 | 0.171 |
| 3 | 1 | 6 | 512 | 128 | 0 | 11 | 0.119 |
| 3 | 1 | 2 | 1024 | 128 | 0 | 49 | 0.180 |
| 3 | 1 | 2 | 256 | 128 | 0 | 19 | 0.122 |
| 3 | 1 | 2 | 128 | 128 | 0 | 23 | 0.126 |
| 3 | 1 | 2 | 64 | 128 | 0 | 14 | 0.099 |
| 3 | 1 | 2 | 1024 | 1024 | 0 | 50 | 0.181 |
| 3 | 1 | 2 | 1024 | 512 | 0 | 25 | 0.168 |
| **3** | **1** | **2** | **1024** | **256** | **0** | **50** | **0.185** |
| 3 | 1 | 2 | 1024 | 128 | 0 | 49 | 0.180 |
| 3 | 1 | 2 | 1024 | 64 | 0 | 48 | 0.179 |
| 3 | 1 | 2 | 1024 | 256 | 0.25 | 38 | 0.145 |
| 3 | 1 | 2 | 1024 | 256 | 0.5 | 11 | 0.084 |
| 3 | 1 | 2 | 1024 | 256 | 0.75 | 12 | 0.081 |

*ii    BERT models*

We fine-tuned all the parameters of the BERT model. The hyperparameters of the $BERT_{BASE}$ and the $BERT_{LARGE}$ models were tuned based on the article's recommendations[8]: 1) batch sizes 16, 32, 64, and 128, 2) different learning rates: $2×10^{-5}$, $3×10^{-5}$, $5×10^{-5}$, and 3) epochs: 2, 3, and 4. The hyperparameters of the remaining models ($BERT_{LARGE}$, $BERT_{MEDIUM}$, $BERT_{SMALL}$, $BERT_{TINY}$) were tuned based on Turc *et al.* recommendations[9]: 1) batch sizes: 8, 16, 32, 64, and 128, 2) learning rates: $1×10^{-4}$, $3×10^{-4}$, $2×10^{-5}$, $3×10^{-5}$, and $5×10^{-5}$, and 3) epochs: 2, 3, and 4. Table S5 to Table S9 show the hyperparameter tuning results for all BERT sizes. **(Table S5-S9)**

**Table S5**: Recall@1 results for hyperparameter tuning of the BERT$_{BASE}$ model on the validation dataset

| Batch size | Learning rate | Epoch | Recall@1 |
|---|---|---|---|
| 16 | $2\times10^{-5}$ | 2 | 0.638 |
| 16 | $2\times10^{-5}$ | 3 | 0.638 |
| 16 | $2\times10^{-5}$ | 4 | 0.631 |
| 16 | $3\times10^{-5}$ | 2 | 0.632 |
| 16 | $3\times10^{-5}$ | 3 | 0.631 |
| 16 | $3\times10^{-5}$ | 4 | 0.627 |
| 16 | $5\times10^{-5}$ | 2 | 0.614 |
| 16 | $5\times10^{-5}$ | 3 | 0.608 |
| 16 | $5\times10^{-5}$ | 4 | 0.611 |
| 32 | $2\times10^{-5}$ | 2 | 0.639 |
| 32 | $2\times10^{-5}$ | 3 | 0.636 |
| 32 | $2\times10^{-5}$ | 4 | 0.631 |
| 32 | $3\times10^{-5}$ | 2 | 0.633 |
| 32 | $3\times10^{-5}$ | 3 | 0.631 |
| 32 | $3\times10^{-5}$ | 4 | 0.627 |
| 32 | $5\times10^{-5}$ | 2 | 0.630 |
| 32 | $5\times10^{-5}$ | 3 | 0.620 |
| 32 | $5\times10^{-5}$ | 4 | 0.616 |
| **64** | **$2\times10^{-5}$** | **2** | **0.641** |
| 64 | $2\times10^{-5}$ | 3 | 0.637 |
| 64 | $2\times10^{-5}$ | 4 | 0.628 |
| 64 | $3\times10^{-5}$ | 2 | 0.639 |
| 64 | $3\times10^{-5}$ | 3 | 0.634 |
| 64 | $3\times10^{-5}$ | 4 | 0.632 |
| 64 | $5\times10^{-5}$ | 2 | 0.639 |
| 64 | $5\times10^{-5}$ | 3 | 0.638 |
| 64 | $5\times10^{-5}$ | 4 | 0.626 |
| 128 | $2\times10^{-5}$ | 2 | 0.639 |
| 128 | $2\times10^{-5}$ | 3 | 0.630 |
| 128 | $2\times10^{-5}$ | 4 | 0.627 |
| 128 | $3\times10^{-5}$ | 2 | 0.631 |
| 128 | $3\times10^{-5}$ | 3 | 0.623 |
| 128 | $3\times10^{-5}$ | 4 | 0.623 |
| 128 | $5\times10^{-5}$ | 2 | 0.630 |
| **128** | $5\times10^{-5}$ | 3 | 0.627 |

**Table S6**: Recall@1 results for hyperparameter tuning of the BERT$_{LARGE}$ on the validation dataset

| Batch size | Learning rate | Epoch | Recall@1 |
| --- | --- | --- | --- |
| 16 | $2\times10^{-5}$ | 2 | 0.061 |
| 16 | $2\times10^{-5}$ | 3 | 0.081 |
| 16 | $2\times10^{-5}$ | 4 | 0.000 |
| 16 | $3\times10^{-5}$ | 2 | 0.097 |
| 16 | $3\times10^{-5}$ | 3 | 0.101 |
| 16 | $3\times10^{-5}$ | 4 | 0.111 |
| 16 | $5\times10^{-5}$ | 2 | 0.085 |
| 16 | $5\times10^{-5}$ | 3 | 0.118 |
| 16 | $5\times10^{-5}$ | 4 | 0.000 |
| **32** | **$2\times10^{-5}$** | **2** | **0.658** |
| 32 | $2\times10^{-5}$ | 3 | 0.644 |
| 32 | $2\times10^{-5}$ | 4 | 0.631 |
| 32 | $3\times10^{-5}$ | 2 | 0.104 |
| 32 | $3\times10^{-5}$ | 3 | 0.108 |
| 32 | $3\times10^{-5}$ | 4 | 0.000 |
| 32 | $5\times10^{-5}$ | 2 | 0.113 |
| 32 | $5\times10^{-5}$ | 3 | 0.003 |
| 32 | $5\times10^{-5}$ | 4 | 0.011 |
| 64 | $2\times10^{-5}$ | 2 | 0.655 |
| 64 | $2\times10^{-5}$ | 3 | 0.647 |
| 64 | $2\times10^{-5}$ | 4 | 0.639 |
| 64 | $3\times10^{-5}$ | 2 | 0.654 |
| 64 | $3\times10^{-5}$ | 3 | 0.645 |
| 64 | $3\times10^{-5}$ | 4 | 0.636 |
| 64 | $5\times10^{-5}$ | 2 | 0.002 |
| 64 | $5\times10^{-5}$ | 3 | 0.103 |
| 64 | $5\times10^{-5}$ | 4 | 0.119 |

**Table S7**: Recall@1 results for hyperparameter tuning of the BERT$_{MEDIUM}$

| Batch size | Learning rate | Epoch | Recall@1 |
|---|---|---|---|
| 8 | $3\times10^{-4}$ | 2 | 0.000 |
| 8 | $3\times10^{-4}$ | 3 | 0.000 |
| 8 | $3\times10^{-4}$ | 4 | 0.000 |
| 8 | $1\times10^{-4}$ | 2 | 0.143 |
| 8 | $1\times10^{-4}$ | 3 | 0.018 |
| 8 | $1\times10^{-4}$ | 4 | 0.018 |
| 8 | $2\times10^{-5}$ | 2 | 0.563 |
| 8 | $2\times10^{-5}$ | 3 | 0.557 |
| 8 | $2\times10^{-5}$ | 4 | 0.557 |
| 8 | $3\times10^{-5}$ | 2 | 0.562 |
| 8 | $3\times10^{-5}$ | 3 | 0.554 |
| 8 | $3\times10^{-5}$ | 4 | 0.554 |
| 8 | $5\times10^{-5}$ | 2 | 0.514 |
| 8 | $5\times10^{-5}$ | 3 | 0.510 |
| 8 | $5\times10^{-5}$ | 4 | 0.510 |
| 16 | $3\times10^{-4}$ | 2 | 0.117 |
| 16 | $3\times10^{-4}$ | 3 | 0.009 |
| 16 | $3\times10^{-4}$ | 4 | 0.009 |
| 16 | $1\times10^{-4}$ | 2 | 0.078 |
| 16 | $1\times10^{-4}$ | 3 | 0.005 |
| 16 | $1\times10^{-4}$ | 4 | 0.005 |
| 16 | $2\times10^{-5}$ | 2 | 0.565 |
| 16 | $2\times10^{-5}$ | 3 | 0.564 |
| 16 | $2\times10^{-5}$ | 4 | 0.564 |
| 16 | $3\times10^{-5}$ | 2 | 0.564 |
| 16 | $3\times10^{-5}$ | 3 | 0.560 |
| 16 | $3\times10^{-5}$ | 4 | 0.560 |
| 16 | $5\times10^{-5}$ | 2 | 0.545 |
| 16 | $5\times10^{-5}$ | 3 | 0.546 |
| 16 | $5\times10^{-5}$ | 4 | 0.546 |
| 32 | $3\times10^{-4}$ | 2 | 0.096 |
| 32 | $3\times10^{-4}$ | 3 | 0.086 |
| 32 | $3\times10^{-4}$ | 4 | 0.086 |
| 32 | $1\times10^{-4}$ | 2 | 0.523 |
| 32 | $1\times10^{-4}$ | 3 | 0.520 |
| 32 | $1\times10^{-4}$ | 4 | 0.520 |
| 32 | $2\times10^{-5}$ | 2 | 0.564 |
| 32 | $2\times10^{-5}$ | 3 | 0.572 |

| Batch size | Learning rate | Epoch | Recall@1 |
|---|---|---|---|
| **32** | **$2 \times 10^{-5}$** | **4** | **0.572** |
| 32 | $3 \times 10^{-5}$ | 2 | 0.568 |
| 32 | $3 \times 10^{-5}$ | 3 | 0.564 |
| 32 | $3 \times 10^{-5}$ | 4 | 0.564 |
| 32 | $5 \times 10^{-5}$ | 2 | 0.555 |
| 32 | $5 \times 10^{-5}$ | 3 | 0.555 |
| 32 | $5 \times 10^{-5}$ | 4 | 0.555 |
| 64 | $3 \times 10^{-4}$ | 2 | 0.100 |
| 64 | $3 \times 10^{-4}$ | 3 | 0.142 |
| 64 | $3 \times 10^{-4}$ | 4 | 0.142 |
| 64 | $1 \times 10^{-4}$ | 2 | 0.546 |
| 64 | $1 \times 10^{-4}$ | 3 | 0.538 |
| 64 | $1 \times 10^{-4}$ | 4 | 0.538 |
| 64 | $2 \times 10^{-5}$ | 2 | 0.566 |
| 64 | $2 \times 10^{-5}$ | 3 | 0.566 |
| 64 | $2 \times 10^{-5}$ | 4 | 0.566 |
| 64 | $3 \times 10^{-5}$ | 2 | 0.565 |
| 64 | $3 \times 10^{-5}$ | 3 | 0.567 |
| 64 | $3 \times 10^{-5}$ | 4 | 0.567 |
| 64 | $5 \times 10^{-5}$ | 2 | 0.559 |
| 64 | $5 \times 10^{-5}$ | 3 | 0.559 |
| 64 | $5 \times 10^{-5}$ | 4 | 0.559 |
| 128 | $3 \times 10^{-4}$ | 2 | 0.003 |
| 128 | $3 \times 10^{-4}$ | 3 | 0.094 |
| 128 | $3 \times 10^{-4}$ | 4 | 0.094 |
| 128 | $1 \times 10^{-4}$ | 2 | 0.562 |
| 128 | $1 \times 10^{-4}$ | 3 | 0.542 |
| 128 | $1 \times 10^{-4}$ | 4 | 0.542 |
| 128 | $2 \times 10^{-5}$ | 2 | 0.571 |
| 128 | $2 \times 10^{-5}$ | 3 | 0.567 |
| 128 | $2 \times 10^{-5}$ | 4 | 0.567 |
| 128 | $3 \times 10^{-5}$ | 2 | 0.570 |
| 128 | $3 \times 10^{-5}$ | 3 | 0.558 |
| 128 | $3 \times 10^{-5}$ | 4 | 0.558 |
| 128 | $5 \times 10^{-5}$ | 2 | 0.566 |
| 128 | $5 \times 10^{-5}$ | 3 | 0.556 |
| 128 | $5 \times 10^{-5}$ | 4 | 0.556 |

**Table S8**: Recall@1 results of the BERT$_{SMALL}$ hyperparameter tuning

| Batch size | Learning rate | Epoch | Recall@1 |
|---|---|---|---|
| 8 | 3x10$^{-4}$ | 2 | 0.090 |
| 8 | 3x10$^{-4}$ | 3 | 0.111 |
| 8 | 3x10$^{-4}$ | 4 | 0.111 |
| 8 | 1x10$^{-4}$ | 2 | 0.458 |
| 8 | 1x10$^{-4}$ | 3 | 0.447 |
| 8 | 1x10$^{-4}$ | 4 | 0.447 |
| 8 | 2x10$^{-5}$ | 2 | 0.540 |
| 8 | 2x10$^{-5}$ | 3 | 0.546 |
| 8 | 2x10$^{-5}$ | 4 | 0.546 |
| 8 | 3x10$^{-5}$ | 2 | 0.535 |
| 8 | 3x10$^{-5}$ | 3 | 0.545 |
| 8 | 3x10$^{-5}$ | 4 | 0.545 |
| 8 | 5x10$^{-5}$ | 2 | 0.481 |
| 8 | 5x10$^{-5}$ | 3 | 0.511 |
| 8 | 5x10$^{-5}$ | 4 | 0.511 |
| 16 | 3x10$^{-4}$ | 2 | 0.094 |
| 16 | 3x10$^{-4}$ | 3 | 0.128 |
| 16 | 3x10$^{-4}$ | 4 | 0.128 |
| 16 | 1x10$^{-4}$ | 2 | 0.495 |
| 16 | 1x10$^{-4}$ | 3 | 0.503 |
| 16 | 1x10$^{-4}$ | 4 | 0.503 |
| 16 | 2x10$^{-5}$ | 2 | 0.544 |
| 16 | 2x10$^{-5}$ | 3 | 0.547 |
| 16 | 2x10$^{-5}$ | 4 | 0.547 |
| 16 | 3x10$^{-5}$ | 2 | 0.536 |
| 16 | 3x10$^{-5}$ | 3 | 0.546 |
| 16 | 3x10$^{-5}$ | 4 | 0.546 |
| 16 | 5x10$^{-5}$ | 2 | 0.541 |
| 16 | 5x10$^{-5}$ | 3 | 0.536 |
| 16 | 5x10$^{-5}$ | 4 | 0.536 |
| 32 | 3x10$^{-4}$ | 2 | 0.107 |
| 32 | 3x10$^{-4}$ | 3 | 0.119 |
| 32 | 3x10$^{-4}$ | 4 | 0.119 |
| 32 | 1x10$^{-4}$ | 2 | 0.512 |
| 32 | 1x10$^{-4}$ | 3 | 0.528 |
| 32 | 1x10$^{-4}$ | 4 | 0.528 |
| 32 | 2x10$^{-5}$ | 2 | 0.541 |
| 32 | 2x10$^{-5}$ | 3 | 0.549 |

| Batch size | Learning rate | Epoch | Recall@1 |
|---|---|---|---|
| 32 | $2\times10^{-5}$ | 4 | 0.549 |
| 32 | $3\times10^{-5}$ | 2 | 0.544 |
| 32 | $3\times10^{-5}$ | 3 | 0.555 |
| **32** | **$3\times10^{-5}$** | **4** | **0.555** |
| 32 | $5\times10^{-5}$ | 2 | 0.534 |
| 32 | $5\times10^{-5}$ | 3 | 0.543 |
| 32 | $5\times10^{-5}$ | 4 | 0.543 |
| 64 | $3\times10^{-4}$ | 2 | 0.389 |
| 64 | $3\times10^{-4}$ | 3 | 0.377 |
| 64 | $3\times10^{-4}$ | 4 | 0.377 |
| 64 | $1\times10^{-4}$ | 2 | 0.526 |
| 64 | $1\times10^{-4}$ | 3 | 0.532 |
| 64 | $1\times10^{-4}$ | 4 | 0.532 |
| 64 | $2\times10^{-5}$ | 2 | 0.535 |
| 64 | $2\times10^{-5}$ | 3 | 0.545 |
| 64 | $2\times10^{-5}$ | 4 | 0.545 |
| 64 | $3\times10^{-5}$ | 2 | 0.544 |
| 64 | $3\times10^{-5}$ | 3 | 0.543 |
| 64 | $3\times10^{-5}$ | 4 | 0.544 |
| 64 | $5\times10^{-5}$ | 2 | 0.535 |
| 64 | $5\times10^{-5}$ | 3 | 0.543 |
| 64 | $5\times10^{-5}$ | 4 | 0.543 |
| 128 | $3\times10^{-4}$ | 2 | 0.427 |
| 128 | $3\times10^{-4}$ | 3 | 0.450 |
| 128 | $3\times10^{-4}$ | 4 | 0.450 |
| 128 | $1\times10^{-4}$ | 2 | 0.540 |
| 128 | $1\times10^{-4}$ | 3 | 0.543 |
| 128 | $1\times10^{-4}$ | 4 | 0.543 |
| 128 | $2\times10^{-5}$ | 2 | 0.530 |
| 128 | $2\times10^{-5}$ | 3 | 0.542 |
| 128 | $2\times10^{-5}$ | 4 | 0.542 |
| 128 | $3\times10^{-5}$ | 2 | 0.530 |
| 128 | $3\times10^{-5}$ | 3 | 0.542 |
| 128 | $3\times10^{-5}$ | 4 | 0.541 |
| 128 | $5\times10^{-5}$ | 2 | 0.546 |
| 128 | $5\times10^{-5}$ | 3 | 0.543 |
| 128 | $5\times10^{-5}$ | 4 | 0.543 |

**Table S9**: Recall@1 results for hyperparameter tuning of the BERT$_{TINY}$

| Batch size | Learning rate | Epoch | Recall@1 |
|---|---|---|---|
| 8 | 3x10$^{-4}$ | 2 | 0.368 |
| 8 | 3x10$^{-4}$ | 3 | 0.389 |
| 8 | 3x10$^{-4}$ | 4 | 0.390 |
| 8 | 1x10$^{-4}$ | 2 | 0.452 |
| 8 | 1x10$^{-4}$ | 3 | 0.467 |
| 8 | 1x10$^{-4}$ | 4 | 0.467 |
| 8 | 2x10$^{-5}$ | 2 | 0.462 |
| 8 | 2x10$^{-5}$ | 3 | 0.467 |
| 8 | 2x10$^{-5}$ | 4 | 0.467 |
| 8 | 3x10$^{-5}$ | 2 | 0.469 |
| 8 | 3x10$^{-5}$ | 3 | 0.474 |
| **8** | **3x10$^{-5}$** | **4** | 0.474 |
| 8 | 5x10$^{-5}$ | 2 | 0.466 |
| 8 | 5x10$^{-5}$ | 3 | 0.471 |
| 8 | 5x10$^{-5}$ | 4 | 0.471 |
| 16 | 3x10$^{-4}$ | 2 | 0.431 |
| 16 | 3x10$^{-4}$ | 3 | 0.427 |
| 16 | 3x10$^{-4}$ | 4 | 0.427 |
| 16 | 1x10$^{-4}$ | 2 | 0.452 |
| 16 | 1x10$^{-4}$ | 3 | 0.461 |
| 16 | 1x10$^{-4}$ | 4 | 0.461 |
| 16 | 2x10$^{-5}$ | 2 | 0.453 |
| 16 | 2x10$^{-5}$ | 3 | 0.455 |
| 16 | 2x10$^{-5}$ | 4 | 0.455 |
| 16 | 3x10$^{-5}$ | 2 | 0.460 |
| 16 | 3x10$^{-5}$ | 3 | 0.460 |
| 16 | 3x10$^{-5}$ | 4 | 0.460 |
| 16 | 5x10$^{-5}$ | 2 | 0.463 |
| 16 | 5x10$^{-5}$ | 3 | 0.464 |
| 16 | 5x10$^{-5}$ | 4 | 0.464 |
| 32 | 3x10$^{-4}$ | 2 | 0.439 |
| 32 | 3x10$^{-4}$ | 3 | 0.432 |
| 32 | 3x10$^{-4}$ | 4 | 0.432 |
| 32 | 1x10$^{-4}$ | 2 | 0.470 |
| 32 | 1x10$^{-4}$ | 3 | 0.460 |
| 32 | 1x10$^{-4}$ | 4 | 0.460 |
| 32 | 2x10$^{-5}$ | 2 | 0.442 |
| 32 | 2x10$^{-5}$ | 3 | 0.451 |

| Batch size | Learning rate | Epoch | Recall@1 |
|---|---|---|---|
| 32 | $2 \times 10^{-5}$ | 4 | 0.451 |
| 32 | $3 \times 10^{-5}$ | 2 | 0.452 |
| 32 | $3 \times 10^{-5}$ | 3 | 0.458 |
| 32 | $3 \times 10^{-5}$ | 4 | 0.458 |
| 32 | $5 \times 10^{-5}$ | 2 | 0.463 |
| 32 | $5 \times 10^{-5}$ | 3 | 0.461 |
| 32 | $5 \times 10^{-5}$ | 4 | 0.461 |
| 64 | $3 \times 10^{-4}$ | 2 | 0.449 |
| 64 | $3 \times 10^{-4}$ | 3 | 0.446 |
| 64 | $3 \times 10^{-4}$ | 4 | 0.446 |
| 64 | $1 \times 10^{-4}$ | 2 | 0.461 |
| 64 | $1 \times 10^{-4}$ | 3 | 0.464 |
| 64 | $1 \times 10^{-4}$ | 4 | 0.464 |
| 64 | $2 \times 10^{-5}$ | 2 | 0.437 |
| 64 | $2 \times 10^{-5}$ | 3 | 0.445 |
| 64 | $2 \times 10^{-5}$ | 4 | 0.445 |
| 64 | $3 \times 10^{-5}$ | 2 | 0.444 |
| 64 | $3 \times 10^{-5}$ | 3 | 0.453 |
| 64 | $3 \times 10^{-5}$ | 4 | 0.453 |
| 64 | $5 \times 10^{-5}$ | 2 | 0.454 |
| 64 | $5 \times 10^{-5}$ | 3 | 0.459 |
| 64 | $5 \times 10^{-5}$ | 4 | 0.459 |
| 128 | $3 \times 10^{-4}$ | 2 | 0.459 |
| 128 | $3 \times 10^{-4}$ | 3 | 0.466 |
| 128 | $3 \times 10^{-4}$ | 4 | 0.466 |
| 128 | $1 \times 10^{-4}$ | 2 | 0.458 |
| 128 | $1 \times 10^{-4}$ | 3 | 0.464 |
| 128 | $1 \times 10^{-4}$ | 4 | 0.464 |
| 128 | $2 \times 10^{-5}$ | 2 | 0.427 |
| 128 | $2 \times 10^{-5}$ | 3 | 0.435 |
| 128 | $2 \times 10^{-5}$ | 4 | 0.435 |
| 128 | $3 \times 10^{-5}$ | 2 | 0.435 |
| 128 | $3 \times 10^{-5}$ | 3 | 0.443 |
| 128 | $3 \times 10^{-5}$ | 4 | 0.443 |
| 128 | $5 \times 10^{-5}$ | 2 | 0.447 |
| 128 | $5 \times 10^{-5}$ | 3 | 0.458 |
| 128 | $5 \times 10^{-5}$ | 4 | 0.458 |

*1.4.   Conversational alignment score calculation*

Each of the 157 dialogues were divided into 10 intervals. Because the transcriptions did not include timestamps, we opted to count the number of tokens and use them to split the dialogue into intervals as evenly as possible without breaking up individual sentences. For instance, in a dialogue with 100 tokens, after a count of 10 tokens a split would be made at the nearest subsequent sentence boundary. Because each dialogue can be tokenized with BPEmb[4] and BERT[8] tokenizers, we may have two different token counts per dialogue.

*i      Mean speaker score:*

In each of the 10 intervals, each response predicted probability (*i*) of a given *speaker* was added up and divided by the total number of responses (*n*) within that interval.

$$Mean\ score_{speaker} = \frac{\sum_i^n g(C,R)_i}{n}$$

*ii      Team difference score:*

By using the mean score of both patients(*p*) and doctors(*d*) per interval (*team* = 2), we calculated the partner similarity. In other words, in each interval *k* we used Litman et al.'s [17] strategy to measure how similar doctor and patient scores were within an interval *k*.

$$TDiff_{k_i} = \frac{\sum_{\forall p \neq d}(|Mean\ score_p - Mean\ score_d|)}{|team| * |team - 1|}$$

*iii      Alignment score:*

After the calculation of partner similarity for every interval *k*, each dialogue ends up with 10 partner similarity scores. We used these to calculate their convergence in any two arbitrary intervals, namely $k_a$ and $k_b$, with the condition that $k_a$ always precedes $s_b$.

$$AS_{(ka,kb)} = TDiff_{ka} - TDiff_{kb} \qquad where\ k_a < k_b \leq 10$$

To avoid deciding which alignment score $AS_{(Sa,Sb)}$ to use for a given dialogue or conversation, we followed Yu et al.'s work[15] and calculated four types of convergence of partner similarity per dialogue or conversation. If any of the scores were not possible to calculate, the overall alignment score was left blank and not analysed. This resulted in the following four types of alignment scores:

$$Max = Max\{AS_{(Sa,Sb)} > 0\} \qquad absMax = Max\{|AS_{(Sa,Sb)}|\}$$

$$Min = Min\{AS_{(Sa,Sb)} > 0\} \qquad absMin = Min\{|AS_{(Sa,Sb)}|\}$$

## 2. Results

*2.1. Comparing using the token [SEP] in modelling context information*

As described in the previous section, the BERT tokenizer adds two additional tokens; the [CLS] token at the beginning of the Context-response (*C,R*) pairs and the separator or [SEP] token in between the context and response, and at the end of response.

Context is derived from concatenated five continuous sentences. We tested whether including a [SEP] after each sentence from the context can have an impact on the performance (i.e., recall@k) of the dialogue modelling. **(Figure S3)**

**Figure S3**: Context with and without the [SEP] token

**1. Context without the [SEP] token**
[CLS] [Sentence_1] [Sentence_2] [Sentence_3] [Sentence_4] [Sentence_5] [SEP] [Response] [SEP]
    ⎵⎯⎯⎯⎯⎯⎯⎯⎯⎯⎯⎯⎯⎯⎯⎯⎯⎯⎯⎯⎯⎯⎯⎯⎯⎵
                        Context

**2. Context with the [SEP] token**
[CLS] [Sentence_1] [SEP] [Sentence_2] [SEP] [Sentence_3] [SEP] [Sentence_4] [SEP] [Sentence_5] [SEP] [Response] [SEP]
    ⎵⎯⎯⎯⎯⎯⎯⎯⎯⎯⎯⎯⎯⎯⎯⎯⎯⎯⎯⎯⎯⎯⎯⎯⎯⎯⎯⎯⎯⎯⎯⎯⎯⎯⎯⎯⎯⎯⎯⎯⎯⎯⎯⎯⎯⎯⎯⎵
                                    Context

The results of testing the inclusion of the [SEP] token in the context from the *C,R* pairs are shown in the **Table S10**.

**Table S10**: Performance of BERT models in dialogue understanding of patient-doctor conversations with and without using the [SEP] token for modelling contextual information

| Model name | Model size | GPUs | Recall@1 | Recall@2 | Recall@5 |
| --- | --- | --- | --- | --- | --- |
| $BERT_{TINY}$ without [SEP] | 4.4M | 1 | 0.471 | 0.625 | 0.856 |
| $BERT_{TINY}$ with [SEP] | 4.4M | 1 | 0.464 | 0.620 | 0.859 |
| $BERT_{SMALL}$ without [SEP] | 28.8M | 1 | 0.552 | 0.711 | 0.908 |
| $BERT_{SMALL}$ with [SEP] | 28.8M | 1 | 0.540 | 0.699 | 0.904 |
| $BERT_{MEDIUM}$ without [SEP] | 41.4M | 1 | 0.568 | 0.731 | 0.915 |
| $BERT_{MEDIUM}$ with [SEP] | 41.4M | 1 | 0.569 | 0.724 | 0.916 |
| $BERT_{BASE}$ without [SEP] | 109M | 4 | 0.640 | 0.792 | 0.942 |
| $BERT_{BASE}$ with [SEP] | 109M | 4 | 0.647 | 0.791 | 0.941 |
| $BERT_{LARGE}$ without [SEP] | 335M | 8 | 0.662 | 0.789 | 0.945 |
| $BERT_{LARGE}$ with [SEP] | 335M | 8 | 0.101 | 0.212 | 0.490 |

## 2.2. Comparing DL with stylebook vs. DL without stylebook

**Table S11**: Random-effects multivariable analysis of Conversational Alignment Scores by Outcome and Model

| Outcome | Model | CA method | Unadjusted analysis Estimate (±SE) | p value | Adjusted analysis* Estimate (±SE) | p value | Mixed-effects† Estimate (±SE) | p value |
|---|---|---|---|---|---|---|---|---|
| OPTION12 | DL without Stylebook | **AbsMax** | **23.05 (±7.98)** | **0.005** | **21.40 (±8.65)** | **0.015** | **18.36 (±7.74)** | **0.020** |
| | | AbsMin | 1.00 (±0.76) | 0.190 | 1.02 (±0.76) | 0.184 | 0.90 (±0.68) | 0.186 |
| | | **Max** | **25.27 (±7.97)** | **0.002** | **22.90 (±8.48)** | **0.008** | **21.01 (±7.63)** | **0.007** |
| | | Min | 10.64 (±44.05) | 0.810 | 18.14 (±43.80) | 0.679 | 7.48 (±38.81) | 0.848 |
| | DL with Stylebook | AbsMax | 16.57 (±6.35) | 0.010 | 15.95 (±6.68) | 0.018 | 8.40 (±6.17) | 0.177 |
| | | AbsMin | 1.71 (±0.74) | 0.023 | 1.67 (±0.75) | 0.028 | 1.11 (±0.75) | 0.141 |
| | | **Max** | **18.79 (±6.30)** | **0.003** | **18.49 (±6.55)** | **0.006** | **10.82 (±6.10)** | **0.080** |
| | | Min | 10.77 (±61.37) | 0.861 | 28.44 (±62.20) | 0.648 | 18.75 (±55.65) | 0.737 |
| DCS | DL without Stylebook | AbsMax | 1.44 (±8.99) | 0.873 | 3.90 (±9.77) | 0.690 | 4.74 (±9.43) | 0.617 |
| | | AbsMin | 0.99 (±0.84) | 0.243 | 0.82 (±0.85) | 0.337 | 0.89 (±0.82) | 0.283 |
| | | Max | 2.41 (±9.00) | 0.789 | 4.94 (±9.57) | 0.606 | 5.39 (±9.28) | 0.563 |
| | | Min | 27.85 (±47.87) | 0.562 | 32.70 (±47.54) | 0.493 | 29.92 (±45.92) | 0.517 |
| | DL with Stylebook | AbsMax | -4.96 (±7.12) | 0.487 | -3.87 (±7.50) | 0.607 | 0.17 (±7.31) | 0.981 |
| | | AbsMin | -0.57 (±0.83) | 0.445 | -0.35 (±0.85) | 0.676 | -0.01 (±0.84) | 0.989 |
| | | Max | -1.28 (±7.05) | 0.856 | 0.32 (±7.35) | 0.965 | 3.52 (±7.17) | 0.625 |
| | | Min | -30.08 (±66.37) | 0.651 | -9.81 (±67.29) | 0.884 | -4.18 (±65.21) | 0.900 |

CA: Conversational alignment. DL: Deep learning
* Adjusted for age, sex, race, and clinical trial arm (intervention vs. control).
† Mixed-effects for clinician and adjusted for age, sex, race, and clinical trial arm (intervention vs. control).

## 2.3. Comparing BERT fine-tuning without fine-tuning

**Table S12**: Random-effects multivariable analysis of Conversational Alignment Scores by Outcome and Model

| Outcome | Model | CA Method | Unadjusted analysis Estimate (±SE) | p value | Adjusted analysis* Estimate (±SE) | p value | Mixed-effects† Estimate (±SE) | p value |
|---|---|---|---|---|---|---|---|---|
| OPTION12 | BERT$_{BASE}$ - 110M with fine-tuning | AbsMax | 0.18 (±11.03) | 0.987 | -2.48 (±11.44) | 0.828 | -3.3 (±10.89) | 0.762 |
| | | AbsMin | -0.27 (±0.86) | 0.753 | 0.33 (±0.89) | 0.712 | 0.08 (±0.83) | 0.920 |
| | | Max | 2.24 (±11.39) | 0.844 | -0.83 (±11.95) | 0.944 | -1.30 (±11.20) | 0.908 |
| | | Min | -61.40 (±59.67) | 0.305 | -49.94 (±59.16) | 0.400 | -38.33 (±48.93) | 0.436 |
| | BERT$_{BASE}$ - 110M without fine-tuning | AbsMax | 13.17 (±8.09) | 0.105 | 14.32 (±8.09) | 0.079 | 9.52 (±7.02) | 0.179 |
| | | AbsMin | -1.42 (±0.99) | 0.155 | -1.20 (±1.02) | 0.241 | -1.41 (±0.92) | 0.131 |
| | | Max | 12.31 (±7.99) | 0.126 | 13.06 (±8.02) | 0.106 | 8.51 (±6.96) | 0.225 |
| | | Min | -28.17 (±81.49) | 0.730 | 23.82 (±82.33) | 0.773 | -7.00 (±77.32) | 0.928 |
| DCS | BERT$_{BASE}$ - 110M with fine-tuning | AbsMax | -18.81 (±11.99) | 0.119 | -23.16 (±12.49) | 0.066 | -21.68 (±12.31) | 0.082 |
| | | AbsMin | -0.17 (±0.96) | 0.855 | 0.33 (±1.00) | 0.735 | 0.49 (±0.98) | 0.618 |
| | | **Max** | **-24.35 (±12.24)** | **0.049** | **-29.75 (±12.85)** | **0.022** | **-27.6 (±12.63)** | **0.032** |
| | | Min | 20.59 (±64.59) | 0.750 | 25.85 (±64.10) | 0.687 | 32.84 (±61.09) | 0.592 |
| | BERT$_{BASE}$ - 110M without fine-tuning | AbsMax | -2.56 (±8.98) | 0.776 | -3.39 (±9.03) | 0.708 | -0.18 (±8.66) | 0.983 |
| | | AbsMin | 0.05 (±1.10) | 0.959 | 0.36 (±1.12) | 0.745 | 0.66 (±1.10) | 0.550 |
| | | Max | -0.56 (±8.84) | 0.949 | -1.00 (±8.92) | 0.911 | 1.75 (±8.54) | 0.838 |
| | | Min | -11.04 (±88.19) | 0.901 | 12.11 (±89.29) | 0.892 | 44.14 (±87.64) | 0.616 |

CA: Conversational alignment
* Adjusted for age, sex, race, and clinical trial arm (intervention vs. control).
† Mixed-effects for clinician and adjusted for age, sex, race, and clinical trial arm (intervention vs. control).

## 2.4. Comparing BERT models with different sizes

**Table S13**: Random-effects multivariable analysis of Conversational Alignment Scores by Outcome and Model

| Outcome | Model | CA Method | Unadjusted Analysis Estimate (±SE) | p value | Adjusted Analysis* Estimate (±SE) | p value | Mixed-effects† Estimate (±SE) | p value |
|---|---|---|---|---|---|---|---|---|
| OPTION12 | BERT$_{LARGE}$ 330M | AbsMax | 5.32 (±10.05) | 0.597 | 4.37 (±10.46) | 0.677 | 5.38 (±9.54) | 0.574 |
| | | AbsMin | 0.59 (±0.93) | 0.524 | 0.67 (±0.93) | 0.471 | 0.87 (±0.85) | 0.311 |
| | | Max | 4.14 (±10.08) | 0.682 | 2.94 (±10.55) | 0.780 | 3.64 (±9.52) | 0.703 |
| | | Min | 32.13 (±97.34) | 0.742 | 60.50 (±99.66) | 0.545 | 17.37 (±103.05) | 0.867 |
| | BERT$_{MEDIUM}$ 55M | AbsMax | 12.79 (±10.60) | 0.229 | 11.86 (±10.80) | 0.274 | 13.53 (±9.94) | 0.962 |
| | | AbsMin | 0.14 (±0.90) | 0.863 | 0.40 (±0.91) | 0.662 | 0.04 (±0.83) | 0.175 |
| | | Max | 11.06 (±10.64) | 0.301 | 9.70 (±10.87) | 0.374 | 8.29 (±9.94) | 0.406 |
| | | Min | 34.41 (±61.71) | 0.578 | 76.84 (±62.37) | 0.220 | 55.94 (±56.82) | 0.328 |
| | BERT$_{SMALL}$ 20M | AbsMax | 2.10 (±11.36) | 0.853 | 3.56 (±11.63) | 0.760 | 2.96 (±10.90) | 0.786 |
| | | AbsMin | 0.20 (±0.90) | 0.825 | 0.78 (±0.91) | 0.392 | 0.22 (±0.84) | 0.787 |
| | | Max | 2.56 (±11.11) | 0.818 | 3.56 (±11.36) | 0.754 | 1.33 (±10.51) | 0.899 |
| | | Min | 8.69 (±85.46) | 0.919 | 16.38 (±84.50) | 0.847 | 6.59 (±76.45) | 0.932 |
| | BERT$_{TINY}$ 11M | AbsMax | -2.94 (±14.62) | 0.841 | -5.53 (±15.30) | 0.718 | -7.29 (±14.24) | 0.610 |
| | | AbsMin | 0.56 (±0.90) | 0.538 | 0.39 (±0.91) | 0.668 | 0.58 (±0.84) | 0.495 |
| | | Max | -3.42 (±14.21) | 0.810 | -5.93 (±14.89) | 0.691 | -8.40 (±13.55) | 0.537 |
| | | Min | 49.27 (±60.97) | 0.420 | 70.13 (±61.01) | 0.252 | 64.46 (±54.45) | 0.240 |
| DCS | BERT$_{LARGE}$ 330M | **AbsMax** | **-18.74 (±11.00)** | **0.090** | **-23.63 (±11.43)** | **0.040** | **-23.86 (±11.12)** | **0.034** |
| | | AbsMin | -0.57 (±1.03) | 0.581 | -0.52 (±1.03) | 0.611 | -0.79 (±1.01) | 0.436 |
| | | **Max** | **-21.03 (±10.94)** | **0.056** | **-27.82 (±11.38)** | **0.015** | **-27.47 (±11.03)** | **0.014** |
| | | Min | -105.46 (±105.72) | 0.320 | -100.41 (±108.47) | 0.356 | -87.11 (±108.62) | 0.424 |
| | BERT$_{MEDIUM}$ 55M | **AbsMax** | **-26.62 (±11.46)** | **0.021** | **-29.91 (±11.72)** | **0.011** | **-29.51 (±11.46)** | **0.011** |
| | | AbsMin | 0.50 (±1.00) | 0.614 | 0.52 (±1.02) | 0.609 | 0.72 (±0.99) | 0.470 |
| | | **Max** | **-26.14 (±11.49)** | **0.024** | **-30.45 (±11.72)** | **0.010** | **-29.39 (±11.43)** | **0.012** |
| | | Min | 58.15 (±66.96) | 0.386 | 82.92 (±68.09) | 0.225 | 94.93 (±65.96) | 0.153 |
| | BERT$_{SMALL}$ 20M | AbsMax | -15.82 (±12.43) | 0.205 | -16.95 (±12.82) | 0.188 | -16.64 (±12.53) | 0.188 |
| | | AbsMin | 0.33 (±1.00) | 0.739 | 0.91 (±1.01) | 0.372 | 1.26 (±0.99) | 0.207 |
| | | Max | -17.67 (±12.13) | 0.147 | -19.34 (±12.45) | 0.122 | -18.61 (±12.11) | 0.128 |
| | | Min | 83.01 (±100.25) | 0.409 | 92.52 (±99.38) | 0.353 | 112.62 (±96.96) | 0.248 |
| | BERT$_{TINY}$ 11M | AbsMax | -18.49 (±15.99) | 0.249 | -21.36 (±16.83) | 0.870 | -17.36 (±16.56) | 0.297 |
| | | AbsMin | -0.02 (±1.01) | 0.977 | 0.01 (±1.02) | 0.990 | -0.25 (±1.00) | 0.802 |
| | | Max | -12.61 (±15.56) | 0.419 | -14.97 (±16.37) | 0.361 | -10.68 (±15.95) | 0.505 |
| | | Min | 32.26 (±66.40) | 0.627 | 55.87 (±66.59) | 0.402 | 66.34 (±64.15) | 0.304 |

CA: Conversational alignment
* Adjusted for age, sex, race, and clinical trial arm (intervention vs. control).
† Mixed-effects for clinician and adjusted for age, sex, race, and clinical trial arm (intervention vs. control).

## 2.5. *Step by step calculation of the conversational alignment score*

**Figure S4**: Conversational alignment score calculation with the BERT model for the conversation AFAP018.

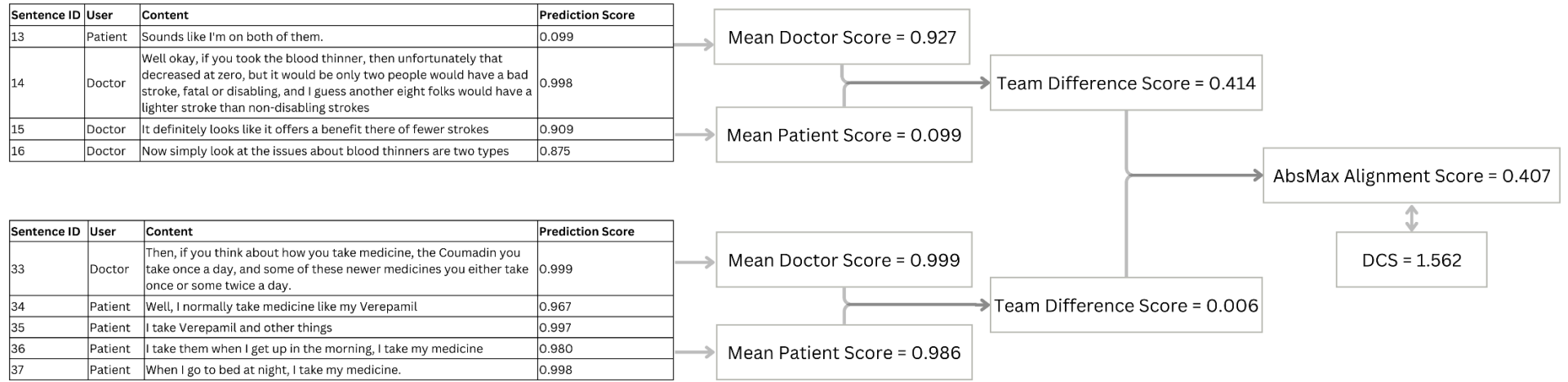

**Figure S5**: Conversational alignment score calculation with the BERT model for the conversation AFPP326.

Conversation: AFPP326 - BERT model

| Sentence ID | User | Content | Prediction Score |
|---|---|---|---|
| 6 | Doctor | All right, This is perfect. | 0.417 |
| 7 | Doctor | Okay, This is kind of a tool that we're using | 0.866 |
| 8 | Doctor | We can kind of go over it | 0.977 |
| 9 | Doctor | I don't know how well you can see | 0.639 |
| 10 | Doctor | Can you see okay here? | 0.949 |
| 11 | Patient | Yeah. | 0.722 |
| 12 | Doctor | All right, What we're gonna do is Rick how old are you again? | 0.660 |
| 13 | Patient | Sixty-three. | 0.991 |
| 14 | Doctor | Sixty-three | 0.997 |
| 15 | Doctor | All right, Basically what we do is we try | 0.838 |
| 16 | Patient | Let me get my glasses on. | 0.209 |
| 17 | Doctor | Oh, okay sure. | 0.867 |
| 18 | Doctor | This is kind of a tool that we use | 0.240 |
| 19 | Doctor | I'm just gonna kinda go over it over. | 0.818 |
| 20 | Patient | All right. | 0.893 |
| 21 | Doctor | We kinda use kinda of this calculator | 0.992 |
| 22 | Doctor | What we're trying to do is what is your five year risk of anything bad happening specifically a stroke? When you have atrial fibrillation you're at risk of developing a stroke | 0.858 |

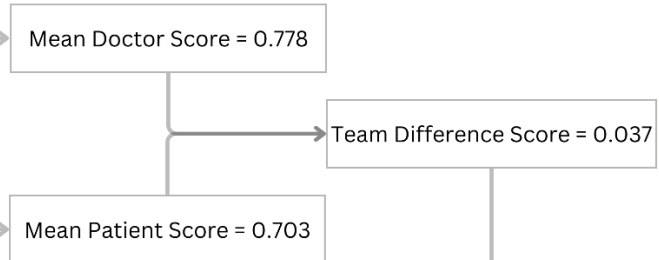

| Sentence ID | User | Content | Prediction Score |
|---|---|---|---|
| 49 | Patient | I didn't say nothing. | 0.917 |
| 50 | Doctor | Okay, What are the benefits with it? If you kinda look at bleeding risk because they're both blood thinners, if you take a blood thinner the benefit the problem is you can bleed more easily, you bruise a little bit more easily | 0.879 |
| 51 | Doctor | In certain situations you could require an emergency treatment | 0.927 |
| 52 | Doctor | The Warfarin that's the the one you're taking right now is you just take it once a day | 0.890 |
| 53 | Doctor | You don't even think about it | 0.959 |
| 54 | Doctor | Don't need any blood tests, right? | 0.910 |
| 55 | Patient | Right. | 0.949 |
| 56 | Doctor | Whereas Warfarin is a little bit different | 0.954 |
| 57 | Doctor | You take it once a day | 0.998 |
| 58 | Doctor | You do need to take blood tests periodically because we don't know sometimes the blood is gonna be too thick, sometimes it's gonna be too thin, all those sorts of things | 0.928 |

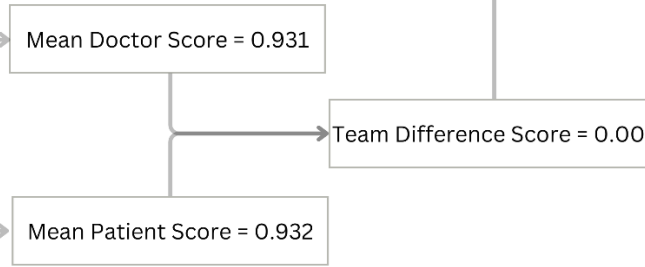

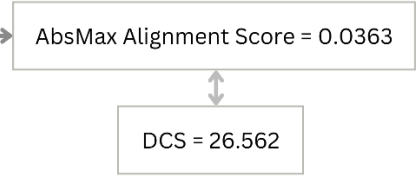